\documentclass[11pt,a4paper]{article}
\usepackage[hyperref]{acl2020}
\usepackage{times}
\usepackage{latexsym}

\usepackage{microtype}

\aclfinalcopy 

\usepackage{array}
\usepackage{calc}
\usepackage{gensymb}
\usepackage{graphicx}
\usepackage{multirow}
\usepackage{textgreek}
\usepackage[caption=false]{subfig}
\usepackage{wrapfig}
\usepackage{sparklines}
\usepackage{colortbl}

\definecolor{noun}{HTML}{882255}
\definecolor{verb}{HTML}{117733}
\definecolor{adj}{HTML}{332288}
\definecolor{adv}{HTML}{CC6677}
\definecolor{other}{HTML}{999933}
\definecolor{ldagray}{HTML}{666666}
\definecolor{bertblue}{HTML}{0072B2}
\definecolor{gptgreen}{HTML}{009E73}
\definecolor{robertared}{HTML}{CC79A7}

\newcommand\wikitext{\textsc{Wikipedia}}
\newcommand{\scotus}{\textsc{Scotus}}
\newcommand{\reviews}{\textsc{Reviews}}
\newcommand{\bert}{BERT}
\newcommand{\roberta}{RoBERTa}
\newcommand{\gpt}{GPT-2}

\newlength{\savedintextsep} 

\title{Topic Modeling with Contextualized Word Representation Clusters}

\author{Laure Thompson \\
  University of Massachusetts Amherst  \\
  \texttt{laurejt@cs.umass.edu} \\\And
  David Mimno \\
  Cornell University \\
  \texttt{mimno@cornell.edu} \\}

\date{}

\begin{document}
\maketitle
\begin{abstract}
Clustering token-level contextualized word representations produces output that shares many similarities with topic models for English text collections.
Unlike clusterings of vocabulary-level word embeddings, the resulting models more naturally capture polysemy and can be used as a way of organizing documents.
We evaluate token clusterings trained from several different output layers of popular contextualized language models.
We find that \bert{} and \gpt{} produce high quality clusterings, but \roberta{} does not.
These cluster models are simple, reliable, and can perform as well as, if not better than, LDA topic models, maintaining high topic quality even when the number of topics is large relative to the size of the local collection.
\end{abstract}

\arrayrulecolor{lightgray}

\section{Introduction}
Contextualized word representations such as those produced by \bert{} \cite{devlin-etal-2019-bert} have revolutionized natural language processing for a number of structured prediction problems.
Recent work has shown that these contextualized representations can support \textit{type-level} semantic clusters \cite{sia2020tired}.
In this work we show that \textit{token-level} clustering provides contextualized semantic information equivalent to that recovered by statistical topic models \cite{blei2003latent}.
From the perspective of contextualized word representations, this result suggests new directions for semantic analysis using both existing models and new architectures more specifically suited for such analysis.
From the perspective of topic modeling, this result implies that transfer learning through contextualized word representations can fill gaps in probabilistic modeling (especially for short documents and small collections) but also suggests new approaches for latent semantic analysis that are more closely tied to mainstream transformer architectures.
 
Topic modeling is often associated with probabilistic generative models in the machine learning literature, but from the perspective of most actual applications the core benefit of such models is that they provide an interpretable latent space that is grounded in the text of a specific collection.
Standard topic modeling algorithms operate by estimating the assignment of individual tokens to topics, either through a Gibbs sampling state or through parameters of variational distributions.
These token-level assignments can then provide disambiguation of tokens based on context, a broad overview of the themes of a corpus, and visualizations of the location of those themes within the corpus \cite{boyd2017applications}.

A related but distinct objective is vocabulary clustering.
These methods operate at the level of distinct word types, but have no inherent connection to words in context \cite[e.g.][]{brown1992classbased,arora2013practical,lancichinetti2015high}.
Recently, there has also been considerable interest in continuous type-level embeddings such as GloVe \cite{pennington-etal-2014-glove} and word2vec \cite{word2veca,word2vecb}, which can be clustered to form interpretable semantic groups.
Although it has not been widely used, the original word2vec distribution includes code for $k$-means clustering of vectors.
\citet{sia2020tired} extends this behavior to contextualized embeddings, but does not take advantage of the contextual, token-based nature of such embeddings.

In this work, we demonstrate a new property of contextualized word representations: if you run a simple $k$-means algorithm on token-level embeddings, the resulting word clusters share similar properties to the output of an LDA model.
Traditional topic modeling can be viewed as token clustering.
Indeed, a clustering of tokens based on \bert{} vectors is functionally indistinguishable from a Gibbs sampling state for LDA, which assigns each token to exactly one topic.
For topic modeling, clustering is based on local context (the current topic disposition of words in the same document) and on global information (the current topic disposition of other words of the same type).
We find that contextualized representations offer similar local and global information, but at a richer and more representationally powerful level.

We argue that pretrained contextualized embeddings provide a simple, reliable method for users to build fine-grained, semantically rich representations of text collections, even with limited local training data.
While for this study we restrict our attention to English text, we see no reason contextualized models trained on non-English data \cite[e.g.][]{martin-etal-2020-camembert,nguyen2020phobert} would not have the same properties.
It is important to note, however, that we make no claim that clustering contextualized word representations is the optimal approach in all or even many situations.
Rather, our goal is to demonstrate the capabilities of contextualized embeddings for token-level semantic clustering and to offer an additional useful application in cases where models like \bert{} are already in use.

\begin{table*}[t]
    \centering
    \small
    \begin{tabular}{|p{.05\textwidth}|p{.065\textwidth}|p{.78\textwidth}|}
    \hline
        Term & Model & Top Words \\
        \hline
        \multirow{6}{*}{land}
        & \multirow{2}{*}{LDA}
        & sea coast Beach Point coastal \textit{\textbf{land}} Long Bay m sand beach tide Norfolk shore Ocean Coast areas \\\cline{3-3}
        & & \textit{\textbf{land}} acres County ha facilities State location property acre cost lot site parking settlers Department \\\cline{2-3}
        & \multirow{2}{*}{\bert{}}
        & arrived arrival landing landed arriving arrive returning settled departed \textit{\textbf{land}} leaving sailed arrives \\\cline{3-3}
        & & \textit{\textbf{land}} property rights estate acres lands territory estates properties farm farmland Land fields acre \\\cline{2-3}
        & \multirow{2}{*}{\gpt{}}
        & arrived landed arriving landfall arrive arrives arrival landing \textit{\textbf{land}} departed ashore embarked Back \\\cline{3-3}
        & & \textit{\textbf{land}} sea ice forest rock mountain ground sand surface beach ocean soil hill lake snow sediment \\
        \hline
        \multirow{6}{*}{metal}
        & \multirow{2}{*}{LDA}
        & metals \textit{\textbf{metal}} potassium sodium + lithium compounds electron ions hydrogen chemical atomic -- ion \\\cline{3-3}
        & & \textit{\textbf{metal}} folk bands music genre band debut Metal heavy musicians lyrics instruments acts groups \\\cline{2-3}
        & \multirow{2}{*}{\bert{}}
        & metals elements \textit{\textbf{metal}} electron element atomic periodic electrons chemical atoms ions atom \\\cline{3-3}
        & & rock dance pop \textit{\textbf{metal}} Rock folk jazz punk comedy Dance heavy funk alternative soul street club \\\cline{2-3}
        & \multirow{2}{*}{\gpt{}}
        & rock pop hop dance \textit{\textbf{metal}} folk hip punk jazz B soul funk alternative rap heavy disco electronic \\\cline{3-3}
        & & plutonium hydrogen carbon sodium potassium \textit{\textbf{metal}} lithium uranium oxygen diamond radioactive \\
    \hline
    \end{tabular}
    \caption{Automatically selected examples of polysemy in contextualized embedding clusters. Clusters containing ``land'' or ``metal'' as top words from \bert{} $L[-1]$, \gpt{} $L[-2]$, and LDA with $K=500$. All models capture multiple senses of the noun ``metal'', but \bert{} and \gpt{} are better than LDA at capturing the syntactic variation of ``land'' as a verb and noun.}
    \label{tab:polysemy}
\end{table*}

\section{Related Work}
We selected three contextualized language models based on their general performance and ease of accessibility to practitioners: \bert{} \cite{devlin-etal-2019-bert}, \gpt{} \cite{radford2019language}, and \roberta{} \cite{liu2019roberta}.
All three use similar Transformer \cite{vaswani2017attention} based architectures, but their objective functions vary in significant ways.
These models are known to encode substantial information about lexical semantics \cite{petroni-etal-2019-language,vulic2020probing}.

Clustering of \textit{vocabulary-level} embeddings has been shown to produce semantically related word clusters \cite{sia2020tired}.
But such embeddings cannot easily account for polysemy or take advantage of local context to disambiguate word senses since each word type is modeled as a single vector.
Since these embeddings are not grounded in specific documents, we cannot directly use them to track the presence of thematic clusters in a particular collection.
In addition, \citet{sia2020tired} find that reweighting their type-level clustering by corpus frequencies is helpful.
In contrast, such frequencies are ``automatically'' accounted for when we operate on the token level.
Similarly, clusterings of {\em sentence-level} embeddings have been shown to produce semantically related document clusters \cite{aharoni-goldberg-2020-unsupervised}.
But such models cannot represent topic mixtures or provide an interpretable word-based representation without additional mapping from clusters to documents to words.
It is widely known that token-level representations of single word types provide contextual disambiguation.
For example, \citet{coenen2019visualizing} show an example distinguishing uses of \textit{die} between the German article, a verb for ``perish'' and a game piece.
We explore this property on the level of whole collections, looking at all word types simultaneously.

There are a number of models that solve the topic model objective directly using contemporary neural network methods \cite[e.g.][]{srivastava2016neural,miao2017discovering,dieng-etal-2020-topic}.
There are also a number of neural models that incorporate topic models to improve performance on a variety of tasks \cite[e.g.][]{chen2016guided,narayan-etal-2018-dont,wang2018reinforced,peinelt-etal-2020-tbert}.
Additionally, \bert{} has been used for word sense disambiguation \cite{wiedemann2019does}.
In contrast, our goal is not to create hybrid or special-purpose models but to show that simple contextualized embedding clusters support token-level topic analysis \textit{in themselves} with no significant additional modeling.
Since our goal is simply to demonstrate this property and not to declare overall ``winners'', we focus on LDA in empirical comparisons because it is the most widely used and straightforward, highlighting the similarities and differences between contextualized embedding clusters and topics.  

\section{Data and Methods}
We use three real-world corpora of varying size, content, and document length: Wikipedia articles (\wikitext{}), Supreme Court of the United States legal opinions (\scotus{}), and Amazon product reviews (\reviews{}).
We select \wikitext{} for its affinity with the training data of the pretrained models.
Because its texts are similar to ones the models have already seen, \wikitext{} is a ``best-case'' scenario for our clustering algorithms.
If a clustering method performs poorly on \wikitext{}, we expect the method to perform poorly in general. 
In contrast, we select \scotus{} and \reviews{} for their content variability.
Legal opinions tend to be long and contain many technical legal terms, while user-generated product reviews tend to be short and highly variable in content and vocabulary.

\begin{table}[h]
    \footnotesize
    \centering
    \small
    \begin{tabular}{ |l|r|r|r| }
    \hline
         Corpus & Docs & Types & Tokens  \\
        \hline
         \wikitext{} & 1.0K & 22K & 1.2M \\
         \scotus{} & 5.3K & 58K & 10.8M \\
         \reviews{} & 100K & 52K & 9.4M \\
    \hline
    \end{tabular}
    \caption{
    Corpus statistics for number of documents, types, and tokens.
    Document and type counts are listed in thousands (K), token counts in millions (M).
    }
    \label{tab:my_label}
\end{table}

\paragraph{\wikitext{}.}
In this collection, documents are Wikipedia articles (excluding headings).
We randomly selected 1,000 articles extracted from the raw/character-level training split of Wikitext-103 \cite{merity2017pointer}.
We largely use the existing tokenization, but recombine internal splits on dot and comma characters but not hyphens so that ``Amazon @.@ com'' becomes ``Amazon.com'', ``1 @,@ 000'' becomes ``1,000'', and ``best @-@ selling'' becomes ``best - selling''.
\setlength{\intextsep}{\savedintextsep}

\paragraph{\scotus{}.}
In this collection, documents are legal opinions from the Supreme Court of the United States filed from 1980 through 2019.%
\footnote{\url{https://www.courtlistener.com/}}
These documents can be very long, but have a regular structure.

\paragraph{\reviews{}.}
In this collection, documents are Amazon product reviews.
For four product categories (Books, Electronics, Movies and TV, CDs and Vinyl), we select 25,000 reviews from category-level dense subsets of Amazon product reviews \cite{he2016ups,mcauley2015image}.

\paragraph{Data Preparation.}
For \scotus{} and \reviews{}, we tokenize documents using the spaCy NLP toolkit.%
\footnote{\url{https://spacy.io/}}
Tokens are case-sensitive non-whitespace character sequences.
For consistency across models, we also delete all control, format, private-use, and surrogate Unicode codepoints since they are internally removed by \bert{}'s tokenizer.
We extract contextualized word representations from \bert{} (cased version), \gpt{}, and \roberta{} using pretrained models available through the huggingface {\tt transformers} library \cite{Wolf2019HuggingFacesTS}.
All methods break low-frequency words into multiple subword tokens:
BERT uses WordPiece \cite{Wu2016GooglesNM}, while \gpt{} and \roberta{} use a byte-level variant of byte pair encoding (BPE) \cite{sennrich-etal-2016-neural}.
For example, the word \textit{disillusioned} is represented by four subtokens ``di -si -llus -ioned'' in \bert{} and by two subtokens ``disillusion -ed'' in \gpt{} and \roberta{}.
One key difference between these tokenizers is that byte-level BPE can encode all inputs, while WordPiece replaces all Unicode codepoints it has not seen in pretraining with the special token \textit{UNK}.
For simplicity, rather than using a sentence splitter we divide documents into the maximum length subtoken blocks.
To make vocabularies comparable across models with different subword tokenization schemes, we reconstitute the original word tokens by averaging the vectors for subword units \cite{bommasani-etal-2020-interpreting}.

\paragraph{Clustering.}
We cluster tokens using spherical $k$-means \cite{dhillon2001concept} with spkm++ initialization \cite{endo2015spherical} because of its simplicity and high-performance, and cosine similarities are commonly used in other embedding contexts.
Although we extract contextualized features for all tokens, prior to clustering we remove frequent words occurring in more than 25\% of documents and rare words occurring in fewer than five documents.
Each clustering is run for 1000 iterations or until convergence.
For LDA, we train models using Mallet \cite{mccallum2002mallet} with hyperparameter optimization occurring every 20 intervals after the first 50.
For each embedding model, we cluster the token vectors extracted from the final layer $L[-1]$, the penultimate layer $L[-2]$, and the antepenultimate layer $L[-3]$.
\citet{vulic2020probing} suggest combining multiple layers, but no combination we tried provided additional benefit for this specific task.
We consider more than the final hidden layer of each model because of the variability in anisotropy across layers \cite{ethayarajh-2019-contextual}.
In a space where any two words have near perfect cosine similarity, clustering will only capture the general word distribution of the corpus.
Since \newcite{ethayarajh-2019-contextual} has shown \gpt{}'s final layer to be extremely anisotropic, we do not expect to produce viable topics in this case.
For each test case, we build ten models each of size $K \in \{50, 100, 500\}$.

\section{Evaluation Metrics}
We evaluate the quality of ``topics'' produced by clustering contextualized word representations with several quantitative measures.
For all models we use hard topic assignments, so each word token has a word type $w_{i}$ and topic assignment $z$.
Note that we use ``topic'' and ``cluster'' interchangeably.

\paragraph{Word Entropy.}
As a proxy for topic specificity, we measure a topic's word diversity using the conditional entropy of word types given a topic: $-\sum_i \Pr(w_i {\mid} z)\log\Pr(w_i {\mid} z)$.
Topics composed of tokens from a small set of types will have low entropy (minimum 0), while topics more evenly spread out across the whole vocabulary will have high entropy (maximum log of vocabulary size; approx. 10 for \wikitext{}).
There is no best fit between quality and specificity, but extreme entropy scores indicate bad topics.
Topics with extremely low entropy are overly specialized, while those with extremely high entropy are overly general.

\paragraph{Coherence.}
We measure the semantic quality of a topic using two word-cooccurrence-based coherence metrics.
These coherence metrics measure whether a topic's words actually occur together.
Internal coherence uses word cooccurrences from the working collection, while external coherence relies on word cooccurrences from a held-out external collection.
The former measures fit to a dataset, while the latter measures generalization.
For internal coherence we use \newcite{mimno-etal-2011-optimizing}'s topic coherence metric, $\sum_i\sum_{j<1}\log\frac{D(w_i,w_j) + \epsilon}{D(w_i)}$, where $D$ refers to the number of documents that contain a word or word-pair.
For external coherence we use \newcite{newman2010evaluating}'s topic coherence metric: $\sum_i\sum_{j<1}\log\frac{\Pr(w_i,w_j) + \epsilon}{\Pr(w_i)\Pr(w_j)}$, where probabilities are estimated from the number of 25-word sliding windows that contain a word or word-pair in an external corpus.
We use the New York Times Annotated Corpus \cite{sandhaus2008new} as our external collection with documents corresponding to articles (headline, article text, and corrected text) tokenized with spaCy.
For both metrics, we use the top 20 words of each topic and set the smoothing factor $\epsilon$ to $10^{-12}$ to reduce penalties for non-cooccurring words \cite{stevens-etal-2012-exploring}.
We ignore words that do not appear in the external corpus and do not consider topics that have fewer than 10 attested words.
These ``skipped'' topics are often an indicator of model failure.
Higher scores are better.

\paragraph{Exclusivity.}
A topic model can attain high coherence by repeating a single high-quality topic multiple times.
To balance this effect, we measure topic diversity using \newcite{bischof2012summarizing}'s word-level exclusivity metric to quantify how exclusive a word $w$ is to a specific topic $z$: $\frac{\Pr(w_i \mid z)}{\sum_{z'}\Pr(w_i \mid z')}$.
A word prevalent in many topics will have a low exclusivity score near 0, while a word occurring in few topics will have a score near 1.
We lift this measure to topics by computing the average exclusivity of each topic's top 20 words.
While higher scores are not inherently better, low scores are indicative of topics with high levels of overlap.

\section{Results}
We evaluate whether contextualized word representation clusters can group together related words, distinguishing distinct uses of the same word based on local context.
Compared to bag-of-words LDA, we expect contextualized embedding clusters to encode more syntactic information.
As we are not doing any kind of fine-tuning, we expect performance to be best on text similar to the pretraining data.
We also expect contextualized embedding clusters to be useful in describing differences between partitions of a working collection.

\begin{figure}[t]
    \centering
    \includegraphics[trim=0.25cm 0 0 0, clip, width=0.48\textwidth]{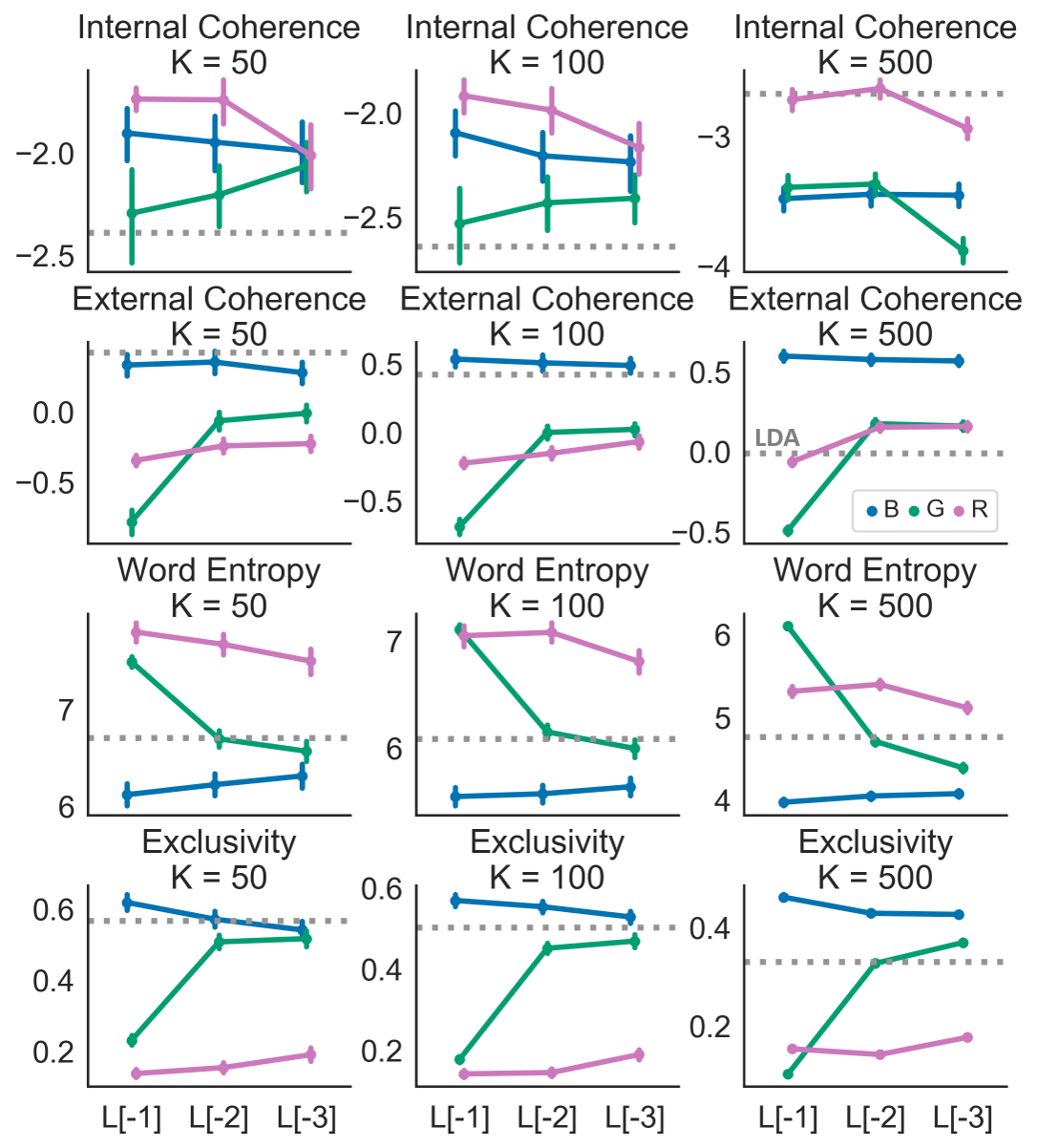}
    \caption{Contextualized embedding clusters produce mean internal and external coherence scores comparable to \textcolor{ldagray}{LDA (dashed line)}. \textcolor{bertblue}{\bert{} clusters (blue)} have high mean external coherence, better than LDA for large numbers of topics. \bert{} clusters contain more unique words, while \textcolor{robertared}{\roberta{} (red)} and \textcolor{gptgreen}{\gpt{} (green)} $L[-1]$ clusters tend to repeat similar clusters. \bert{} clusters have the highest word concentrations.}
    \label{fig:initial_results}
\end{figure}

\begin{figure}[t]
    \centering
    \includegraphics[height=2.5cm]{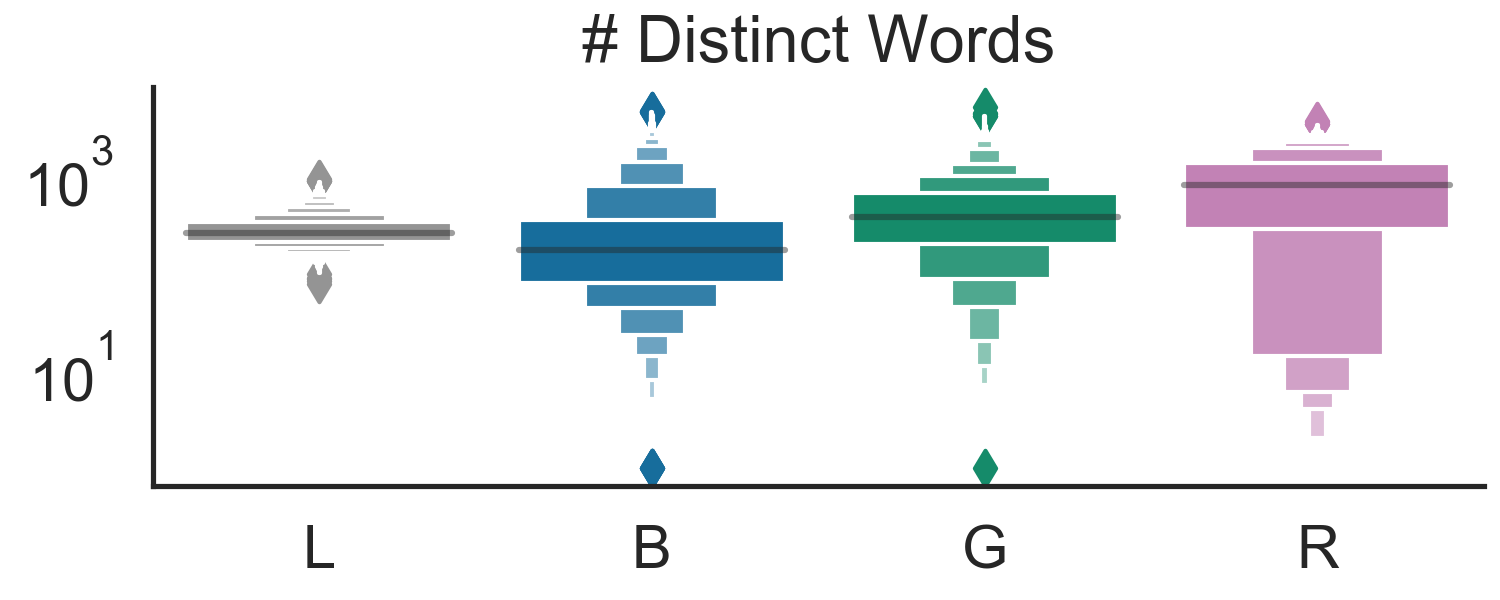}
    \caption{Distinct words per cluster for \textcolor{ldagray}{LDA}, \mbox{\textcolor{bertblue}{\bert{} $L[-1]$}}, \textcolor{gptgreen}{\gpt{} $L[-2]$}, and \textcolor{robertared}{\roberta{} $L[-1]$} for \mbox{$K=500$}. Although \bert{} clusters cover fewer word types on average, \roberta{} produces more clusters with very few ($<20$) word types. 
    }
  \label{fig:n_types}
\end{figure}

\paragraph{\bert{} produces meaningful topic models.}
\bert{} cluster models consistently form semantically meaningful topics, with the final layer performing marginally better for larger $K$.
Figure \ref{fig:initial_results} shows that \bert{} clusterings have the highest external coherence, matching LDA for $K \in \{50, 100\}$ and beating LDA for $K=500$.
For internal coherence, the opposite is true, with \bert{} on par with LDA for smaller $K$, while LDA ``fits'' better for $K=500$.
This distinction suggests that at very fine-grained topics, LDA may be overfitting to noise.
\bert{} has relatively low word entropy, indicating more focused topics on average.
Figure \ref{fig:n_types} shows the number of word types per cluster.
\bert{} clusters are on average smaller than LDA topics (counted from an unsmoothed sampling state), but very few \bert{} clusters fall below our 10-valid-words threshold for coherence scoring.
\bert{} clusters are not only semantically meaningful, but also unique.
Figure \ref{fig:initial_results} shows that \bert{} clusters have exclusivity scores as high if not higher than LDA topics on average.
Since there is little difference between layers, we will only consider \bert{} L[-1] for the remainder of this work.

\paragraph{\gpt{} \textit{can} produce meaningful topic models.}
As expected, the final layer clusterings of \gpt{} form bad topics.
These clusters tend to be homogeneous (low word entropy) and similar to each other (low exclusivity).
They also highlight the differences between our two coherence scores.
Since these clusters tend to repeatedly echo the background distribution of \wikitext{}, they perform relatively well for internal coherence, but poorly for external coherence.
Since the final layer of \gpt{} has such high anisotropy, we cannot expect vector directionalities to encode semantically meaningful information.
In contrast, the penultimate and antepenultimate layer clusterings perform much better.
We see a large improvement in external coherence surpassing LDA for $K=500$.
Topic word entropy and exclusivity are also improved.

For $K=500$, \gpt{} L[-3] has surprisingly low mean internal coherence---the worst scores in Figure \ref{fig:initial_results} by a significant  margin.
The number of topics below the 10-valid-words threshold is similar to \bert{}, so this result is comparable.
We posit that this layer is relying more on transferred knowledge from the pretrained \gpt{} model than the working collection.
Because of this less explained behavior, we will only consider \gpt{} L[-2] going forward.

\paragraph{\roberta{} clusters are noticeably worse.}
Given \bert{}'s success and \gpt{}'s partial success, we were surprised to find that \roberta{} cluster models were consistently of poor quality, with very low exclusivity scores and high word entropies.
Although \roberta{} scores fairly well in coherence, this is not indicative of collectively high quality topics because of the correspondingly low exclusivity scores.
As shown in Figure \ref{fig:initial_results}, \roberta{} has the highest average number of distinct words per cluster, but also large numbers of clusters that contain very few distinct words.
For $K=500$, 25--50 clusters are skipped on average for different layer choices.
For example, one topic consists entirely of the words \textit{game, games, Game}, another just \textit{ago}, and one simply the \textbf{} symbol.
The remaining tokens are thus limited to a smaller number of more general topics that are  closer to the corpus distribution.

While it is commonly accepted that \roberta{} outperforms \bert{} for a variety of natural language understanding tasks \cite{wang2019superglue}, we find the opposite to be true for semantic clustering.
There are a number of differences between \bert{} and \roberta{}, but our experimental results do not mark a clear cause.
The tokenization method is a very unlikely source since \gpt{} uses the same scheme.

\begin{table*}[t]
    \centering
    \small
    \begin{tabular}{|l|l|l|p{.65\textwidth}|}
    \hline
        Model & Perc. & Entr. & Top Words (\textcolor{noun}{noun} \textcolor{verb}{verb} \textcolor{adj}{adj} \textcolor{adv}{adv} \textcolor{other}{other})\\
        \hline
        \multirow{5}{*}{LDA}
& 5\% & 0.69 & \textcolor{noun}{Valley} \textcolor{noun}{Death} \textcolor{noun}{valley} \textcolor{noun}{Creek} \textcolor{noun}{California} \textcolor{noun}{mining} \textcolor{noun}{\degree} \textcolor{noun}{Range} \textcolor{noun}{Nevada} \textcolor{noun}{Desert} \\
& 25\% & 0.97 & \textcolor{noun}{army} \textcolor{noun}{forces} \textcolor{noun}{soldiers} \textcolor{noun}{campaign} \textcolor{noun}{troops} \textcolor{verb}{captured} \textcolor{verb}{defeated} \textcolor{noun}{Battle} \textcolor{noun}{victory} \textcolor{noun}{commander} \\
& 50\% & 1.11 & \textcolor{noun}{society} \textcolor{noun}{News} \textcolor{noun}{Week} \textcolor{adj}{Good} \textcolor{noun}{Spirit} \textcolor{noun}{Fruit} \textcolor{verb}{says} \textcolor{noun}{Doug} \textcolor{noun}{host} \textcolor{adj}{free} \\
& 75\% & 1.28 & \textcolor{noun}{Washington} \textcolor{noun}{Delaware} \textcolor{noun}{ceremony} \textcolor{noun}{Grand} \textcolor{noun}{Capitol} \textcolor{verb}{building} \textcolor{other}{156} \textcolor{noun}{Number} \textcolor{verb}{laying} \textcolor{noun}{Master} \\
& 95\% & 1.53 & \textcolor{noun}{critics} \textcolor{noun}{reviews} \textcolor{noun}{review} \textcolor{adj}{positive} \textcolor{adj}{mixed} \textcolor{noun}{list} \textcolor{noun}{Entertainment} \textcolor{noun}{Times} \textcolor{noun}{style} \textcolor{other}{something} \\
\hline
        \multirow{5}{*}{\bert}
        & 5\% & 0.00 & \textcolor{other}{1997} \textcolor{other}{1996} \textcolor{other}{1995} \textcolor{other}{1937} \textcolor{other}{1895} \textcolor{other}{1935} \textcolor{other}{96} \textcolor{other}{1896} \textcolor{other}{1795} \textcolor{other}{97} \\
& 25\% & 0.61 & \textcolor{adj}{Jewish} \textcolor{noun}{Israel} \textcolor{noun}{Jews} \textcolor{noun}{Ottoman} \textcolor{noun}{Arab} \textcolor{adj}{Muslim} \textcolor{adj}{Israeli} \textcolor{adj}{Islamic} \textcolor{noun}{Jerusalem} \textcolor{noun}{Islam} \\
& 50\% & 0.86 & \textcolor{verb}{captured} \textcolor{verb}{defeated} \textcolor{verb}{attacked} \textcolor{noun}{capture} \textcolor{noun}{attack} \textcolor{noun}{siege} \textcolor{verb}{destroyed} \textcolor{verb}{surrender} \textcolor{noun}{defeat} \textcolor{verb}{occupied} \\
& 75\% & 1.09 & \textcolor{noun}{hop} \textcolor{noun}{dance} \textcolor{noun}{hip} \textcolor{other}{B} \textcolor{noun}{R} \textcolor{noun}{Dance} \textcolor{noun}{Hip} \textcolor{other}{Z} \textcolor{noun}{Hop} \textcolor{noun}{rapper} \\
& 95\% & 1.48 & \textcolor{adj}{separate} \textcolor{verb}{combined} \textcolor{other}{co} \textcolor{adj}{joint} \textcolor{verb}{shared} \textcolor{verb}{divided} \textcolor{adj}{common} \textcolor{noun}{combination} \textcolor{adj}{distinct} \textcolor{adj}{respective} \\
    \hline
    \multirow{5}{*}{\gpt{}}
    & 5\% & 0.00 & \textcolor{other}{2004} \textcolor{other}{2003} \textcolor{other}{2015} \textcolor{other}{2000} \textcolor{other}{2014} \textcolor{other}{1998} \textcolor{other}{2001} \textcolor{other}{2013} \textcolor{other}{2002} \textcolor{other}{1997} \\
& 25\% & 0.42 & \textcolor{noun}{Atlantic} \textcolor{noun}{Pacific} \textcolor{noun}{Gulf} \textcolor{noun}{Mediterranean} \textcolor{noun}{Caribbean} \textcolor{noun}{Columbia} \textcolor{noun}{Indian} \textcolor{noun}{Baltic} \textcolor{noun}{Bay} \textcolor{noun}{Florida} \\
& 50\% & 0.73 & \textcolor{verb}{knew} \textcolor{verb}{finds} \textcolor{noun}{discovers} \textcolor{verb}{learned} \textcolor{noun}{reveals} \textcolor{verb}{discovered} \textcolor{verb}{know} \textcolor{verb}{heard} \textcolor{noun}{discover} \textcolor{noun}{learns} \\
& 75\% & 1.02 & \textcolor{noun}{Olympic} \textcolor{noun}{League} \textcolor{noun}{FA} \textcolor{noun}{Summer} \textcolor{noun}{Premier} \textcolor{noun}{Division} \textcolor{noun}{UEFA} \textcolor{adj}{European} \textcolor{noun}{Winter} \textcolor{verb}{Tour} \\
& 95\% & 1.42 & \textcolor{adj}{positive} \textcolor{adj}{mixed} \textcolor{adj}{critical} \textcolor{adj}{negative} \textcolor{verb}{garnered} \textcolor{adj}{favorable} \textcolor{adv}{mostly} \textcolor{verb}{attracted} \textcolor{noun}{commercial} \\

    \hline
    \end{tabular}
    \caption{Contextualized embedding clusters are more syntactically aware than LDA. Topics ranked by the entropy of POS distribution of the top 20 words (10 shown) with $K=500$. }
    \label{tab:postopics}
\end{table*}

\paragraph{Contextualized embedding clusters capture polysemy.}
A limitation of many methods that rely on vocabulary-level embeddings is that they cannot explicitly account for polysemy and contextual differences in meaning.
In contrast, token-based topic models are able to distinguish differences in meaning between contexts.
There has already been evidence that token-level contextualized embeddings are able to distinguish contextual meanings for specific examples \cite{wiedemann2019does,coenen2019visualizing}, but can they also do this for entire collections?

Instead of manually selecting terms we expect to be polysemous, we choose terms that occur as top words for clusters with dissimilar word distributions (high Jensen-Shannon divergence).
While dissimilarity is not indicative of polysemy---different topics can use a term in the same way---it narrows our focus to words that are more likely to be polysemous. 
In Table \ref{tab:polysemy}, we see topics for two such terms ``land'' and ``metal''.
All models are able to distinguish \textit{metal} the material from \textit{metal} the genre, but \bert{} and \gpt{} are also able to distinguish \textit{land} the noun from \textit{land} the verb.

\paragraph{Contextualized embedding clusters are more syntactically consistent than LDA topics.}
Contextualized word representations are known to represent a large amount of syntactic information that is not available to traditional bag-of-words topic models \cite{Goldberg2019AssessingBS}.
We therefore expect that token-level clusterings of contextualized word representations will have more homogeneity of syntactic form, and indeed we find that they do.

As a simple proxy for syntactic similarity, we find the most likely part of speech (POS) for the top words in each cluster.
We use this method because it is easily implemented; inaccuracies should be consistent across models.
To quantitatively evaluate the homogeneity of POS within each topic, we count the distribution of POS tags for the top 20 words of a cluster and calculate the entropy of that distribution.
If all 20 words are the same POS, this value will be 0, while if POS tags are more diffuse it will be larger.
We find that \bert{} and \gpt{} clusters have consistently lower entropy.
In Table \ref{tab:postopics}, we see that the 25th percentile for LDA topics has entropy 0.97, higher than the median entropy for both \bert{} and \gpt{}.
We find that these results are consistent across model sizes.
Although contextualized embedding clusters are more homogeneous in POS, LDA may appear more homogeneous because it is dominated by nouns.
For LDA, nouns and proper nouns account for 43.7\% and 33.4\% respectively of all the words in the top 20 for all topics, while verbs make up 8.5\% and adjectives 6.7\%.
These proportions are 39.0\%, 25.3\%, 14.9\%, and 8.1\% for \bert{}, and 37.0\%, 23.4\%, 16.9\%, and 9.5\% for \gpt{}.

\paragraph{Compression improves efficiency while maintaining quality.}
We have established that we can effectively learn topic models by clustering token-level contextualized embeddings, and we have shown that there are advantages to clustering at the token rather than vocabulary level.
But for token-level clustering to be more than a curiosity we need to address computational complexity.
Vocabularies are typically on the order of tens of thousands of words, while collections often contain millions of words.
Even storing full 768-dimensional vectors for millions of tokens, much less clustering them, can be beyond the capability of many potential users' computing resources.
Therefore, we investigate the effects of feature dimensionality reduction to reduce the memory footprint of our method.

The hidden layers of deep learning models are known to learn representations with dimensionalities much lower than the number of neurons \cite{raghu2017svcca}.
We apply two methods for dimensionality reduction: principal component analysis (PCA) and sparse random projection (SRP) \cite{Li2006VerySR,Achlioptas2001DatabasefriendlyRP}.\footnote{All implementations from \cite{scikit-learn}.}

\begin{figure}[t]
    \centering
    \captionsetup[subfloat]{farskip=1pt}
    \subfloat{
        \includegraphics[trim=0.25cm 0 0 0, clip, width=0.47\textwidth]{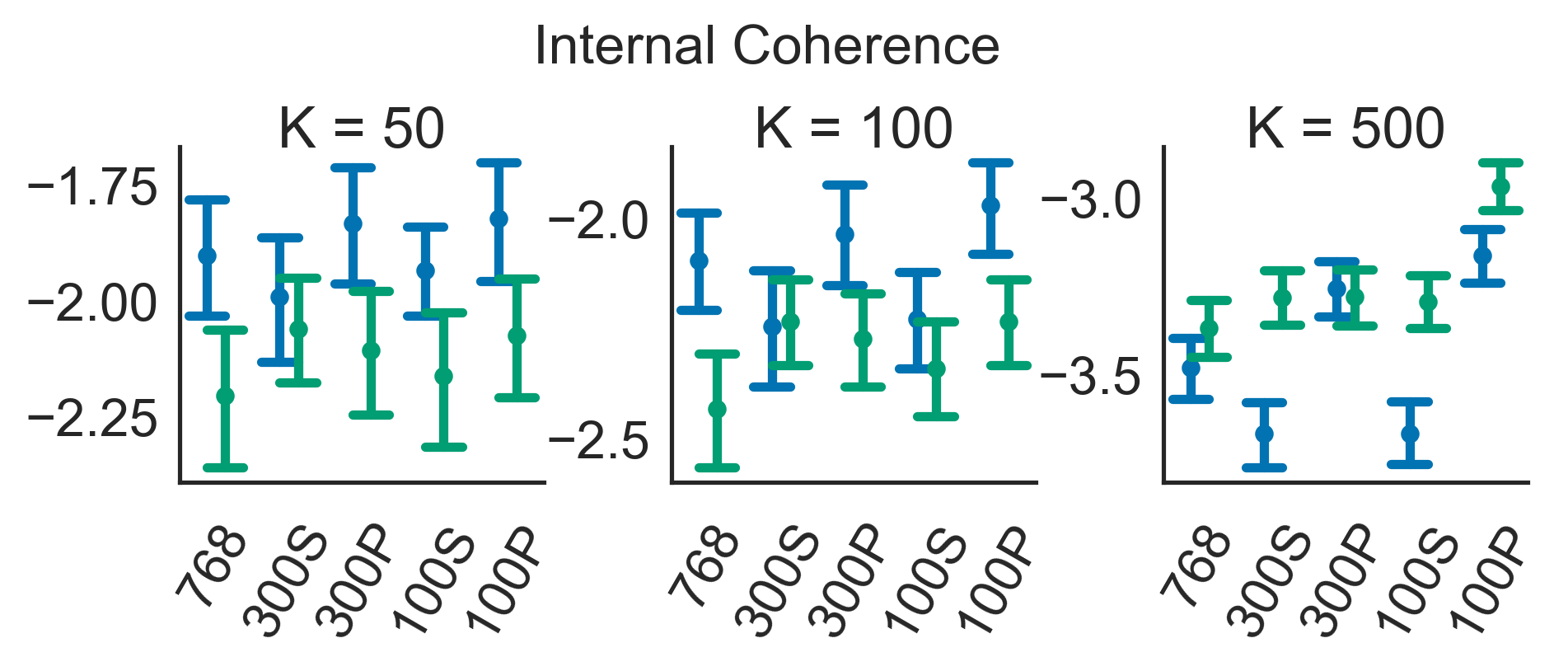}
    }\\
    \subfloat{
        \includegraphics[trim=0.25cm 0 0 0, clip, width=0.47\textwidth]{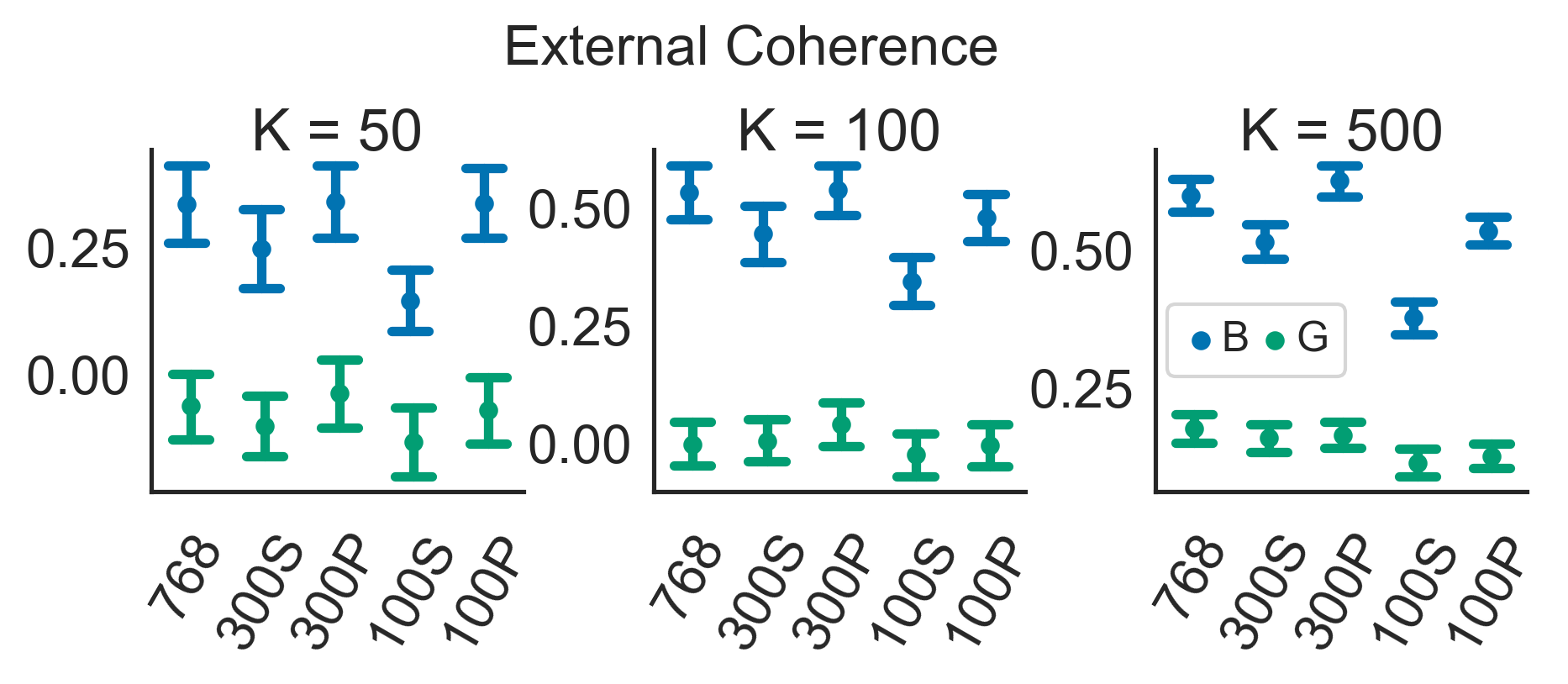}
    }
    \caption{Mean internal and external coherence for reduced features of \textcolor{bertblue}{\bert{}} and \textcolor{gptgreen}{\gpt{}}. Features reduced with PCA tend to have higher coherence than SRP.}
    \label{fig:dims_coherence}
\end{figure}

We find that reducing our token vectors to as few as 100 dimensions can have little negative effect.
Figure \ref{fig:dims_coherence} shows that reduced PCA features produce improved internal coherence and little significant change in external coherence, but reduced SRP features are worse in both metrics, especially for \bert{}.
We note that more clusters pass the 10-valid-words threshold for reduced PCA and SRP features.
Instead of skipping 10 \bert{} clusters and 7 \gpt{} clusters on average, no clusters are skipped for PCA reduced features and only 1 for 100-dimensional SRP features.
For 300-dimensional SRP features there is only a significant drop for \gpt{} with 3 skipped on average.
This decrease in skipped topics indicates that overly specific topics are being replaced with less specific ones.
We hypothesize that dimensionality reduction is smoothing away ``spikes'' in the embeddings space that cause the algorithm to identify small clusters.
Finally, larger dimensionality reductions decrease concentration and exclusivity, making clusters more general.

PCA is significantly better than SRP, especially for more aggressive dimensionality reductions.
We find that mean-centering SRP features does not significantly improve results.
An advantage of SRP, however, is that the projection matrix can be generated offline and immediately applied to embedding vectors as soon as they are generated.
To overcome the memory limitations of PCA, we use a batch approximation, incremental PCA \cite{ross2008incremental}.
Using 100 dimensions and scikit-learn's default batch size of five times the number of features (3840), we find no significant difference in results between PCA and incremental PCA.
For the remainder of experiments we use 100-dimensional vectors which correspond to the top 100 components produced by incremental PCA.

\begin{figure}[t]
    \centering
    \includegraphics[trim=0.25cm 0 0 0, clip, width=0.48\textwidth]{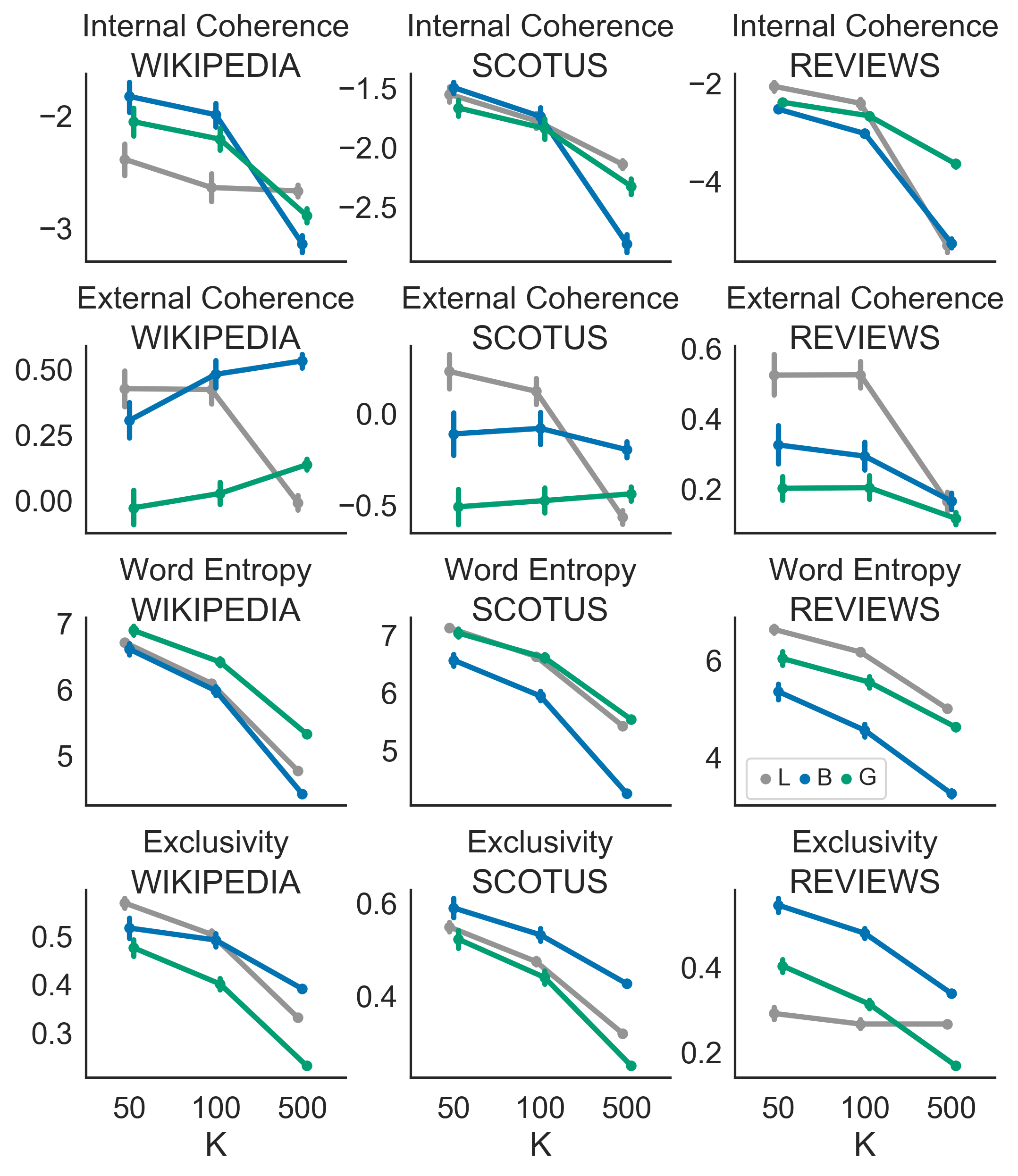}
    \caption{
    \textcolor{bertblue}{\bert{}} and \textcolor{gptgreen}{\gpt{}} produce coherent topics for less familiar (w.r.t. pretraining) collections.
    \textcolor{bertblue}{\bert{}} consistently produces more unique clusters.
    \textcolor{ldagray}{LDA} external coherence drops for $K=500$.}
    \label{fig:collections_coherence}
\end{figure}

\begin{table*}[t]
    \centering
    \small
    \begin{tabular}{|l|p{0.825\textwidth}|}
    \hline
\multirow{3}{*}{books}
& book books author novel novels work Book fiction by authors Kindle published volume works literature \\ \cline{2-2}
& read reading copy Read reads Reading readable reader reread across follow opened researched study hand \\ \cline{2-2}
& problem children problems course power lives mystery questions issues words death example reality battle \\
\hline
\multirow{3}{*}{electronics}
& use up than off used back over using there about  work need down thing nice other full no easy small \\ \cline{2-2}
& screen quality sound device power battery unit system software remote video player mode drive audio \\ \cline{2-2}
& setup remote battery mode card set range input signal support setting manual stand menu GPS fan power \\
\hline
\multirow{3}{*}{movies}
& movie movies films flick theater Movie flicks game cinema film comedies Movies pictures westerns \\ \cline{2-2}
& into up over through between off down than about around during against under found away along though \\ \cline{2-2}
& film movie picture screen documentary films Film cinema feature work production filmmaker piece Picture \\
\hline
\multirow{3}{*}{cds}
& album albums record release Album LP releases records effort up label EP thing titled studio label titled \\ \cline{2-2}
& songs tracks hits tunes singles material stuff Songs ballads cuts ones numbers sounds artists Hits versions \\ \cline{2-2}
& lyrics guitar vocals voice bass singing solo vocal sound work music piano chorus style here live playing \\
\hline
    \end{tabular}
    \caption{Unlike vocabulary-level clusters, token-level clusters are grounded in specific documents and can be used to analyze collections. Here we show the most prominent \bert{} topics ($K=500$) for the four product categories in \reviews{}. This analysis is purely \textit{post hoc}, neither \bert{} nor its clustering have access to product labels.}
    \label{tab:reviewtopics}
\end{table*}

\begin{table*}[h]
\centering
\small
\begin{tabular}{|p{0.94\textwidth}|}
\hline
 than beyond Than twice upon much except besides half times less unlike nor yet per alone exceeded beside above within \\
\hline
 day days Day morning today date daily Days month 19 basis night period 1979 shortly moment term rainy mood sunny \\
\hline
 can Can able possible manage could knows lets capable allows can't easily s Cannot cant ability Can't can`t barely cannot \\
\hline
 when When once time whenever Once soon everytime upon during Whenever moment Everytime At near before anytime \\
\hline
 too enough Too overly taste beyond tired sufficiently plenty somehow Enough unnecessary sufficient sick half overkill  \\
\hline
 because since due Because Since considering cause meaning given thanks means based result order therefore being \\
\hline
 went took got happened came started did turned ended fell used kept left taken gave stopped won ran made moved \\
\hline
 would 'd Would might d normally I'd imagine otherwise happily woulda wouldn't Wouldn't envision probably Iwould \\
\hline
 up off Up ready upload ups along end forth uploading used down away unpacked securely rope onto open unpacking \\
\hline
 instead rather either matter Instead opposed other depending based choice Rather than favor otherwise regardless no \\
 \hline
    \end{tabular}
    \caption{The ten \bert{} topics from \reviews{} with the most \textit{uniform} distribution over product categories.}
    \label{tab:eventopics}
\end{table*}

\paragraph{Pretrained embeddings are effective for different collections.}
While \bert{} and \gpt{} cluster models produce useful topics for \wikitext{}, will this hold for collections less similar to the training data of these pretrained models?
Does it work well for collections of much shorter and much longer texts than Wikipedia articles?
We find that both \bert{} and \gpt{} produce semantically meaningful topics for \scotus{} and \reviews{}, but \bert{} continues to outperform \gpt{}.
As with \wikitext{}, we find that contextualized embedding clusters have the largest advantage over LDA for large $K$.
Figure \ref{fig:collections_coherence} shows that for $K=500$ \bert{} and \gpt{} clusters have significantly higher external coherence scores on average than LDA topics for \scotus{} and very similar scores for \reviews{}.
For smaller $K$, LDA has the highest external coherence scores followed by \bert{}.
Internal coherence is more difficult to interpret because of the variability in exclusivity.
With $K=500$, \bert{} clusters have substantially worse internal coherence scores. 
In contrast, \gpt{} clusters tend to experience a smaller drop in scores, but this can be partially explained by their much lower average exclusivity.
We find that \bert{} consistently produces the most unique topics for \scotus{} and \reviews{}.
\bert{} consistently has significantly higher mean exclusivity scores for both \scotus{} and \reviews{}, while \gpt{} tends to have scores as good as LDA for $K\in \{50,100\}$, but significantly lower for $K=500$.

\begin{table*}[h]
    \centering
    \small
    \begin{tabular}{|p{0.1\textwidth}|p{0.75\textwidth}|}
    \hline
    1980--2019 & Top Words \\
\hline
\begin{sparkline}{10}
\sparkspike 0.000 0.495 \sparkspike 0.026 1.000 \sparkspike 0.051 0.962 \sparkspike 0.077 0.937 \sparkspike 0.103 0.700 \sparkspike 0.128 0.637 \sparkspike 0.154 0.643 \sparkspike 0.179 0.734 \sparkspike 0.205 0.575 \sparkspike 0.231 0.807 \sparkspike 0.256 0.484 \sparkspike 0.282 0.557 \sparkspike 0.308 0.147 \sparkspike 0.333 0.170 \sparkspike 0.359 0.282 \sparkspike 0.385 0.204 \sparkspike 0.410 0.245 \sparkspike 0.436 0.203 \sparkspike 0.462 0.321 \sparkspike 0.487 0.240 \sparkspike 0.513 0.080 \sparkspike 0.538 0.153 \sparkspike 0.564 0.212 \sparkspike 0.590 0.204 \sparkspike 0.615 0.117 \sparkspike 0.641 0.107 \sparkspike 0.667 0.119 \sparkspike 0.692 0.142 \sparkspike 0.718 0.185 \sparkspike 0.744 0.222 \sparkspike 0.769 0.147 \sparkspike 0.795 0.240 \sparkspike 0.821 0.234 \sparkspike 0.846 0.160 \sparkspike 0.872 0.264 \sparkspike 0.897 0.159 \sparkspike 0.923 0.095 \sparkspike 0.949 0.043 \sparkspike 0.974 0.260 \sparkspike 1.000 0.152
\end{sparkline} & union employment labor bargaining Labor workers job strike unions working \\
\hline
\begin{sparkline}{10}
\sparkspike 0.000 0.515 \sparkspike 0.026 1.000 \sparkspike 0.051 0.442 \sparkspike 0.077 0.918 \sparkspike 0.103 0.390 \sparkspike 0.128 0.358 \sparkspike 0.154 0.309 \sparkspike 0.179 0.391 \sparkspike 0.205 0.400 \sparkspike 0.231 0.383 \sparkspike 0.256 0.126 \sparkspike 0.282 0.171 \sparkspike 0.308 0.202 \sparkspike 0.333 0.169 \sparkspike 0.359 0.154 \sparkspike 0.385 0.129 \sparkspike 0.410 0.032 \sparkspike 0.436 0.290 \sparkspike 0.462 0.248 \sparkspike 0.487 0.238 \sparkspike 0.513 0.161 \sparkspike 0.538 0.078 \sparkspike 0.564 0.104 \sparkspike 0.590 0.170 \sparkspike 0.615 0.159 \sparkspike 0.641 0.177 \sparkspike 0.667 0.053 \sparkspike 0.692 0.057 \sparkspike 0.718 0.105 \sparkspike 0.744 0.237 \sparkspike 0.769 0.121 \sparkspike 0.795 0.065 \sparkspike 0.821 0.079 \sparkspike 0.846 0.077 \sparkspike 0.872 0.054 \sparkspike 0.897 0.197 \sparkspike 0.923 0.049 \sparkspike 0.949 0.009 \sparkspike 0.974 0.062 \sparkspike 1.000 0.426
\end{sparkline} & gas coal oil natural mining mineral fuel mine fishing hunting \\
\hline
\begin{sparkline}{10}
\sparkspike 0.000 0.739 \sparkspike 0.026 1.000 \sparkspike 0.051 0.550 \sparkspike 0.077 0.749 \sparkspike 0.103 0.536 \sparkspike 0.128 0.492 \sparkspike 0.154 0.461 \sparkspike 0.179 0.435 \sparkspike 0.205 0.505 \sparkspike 0.231 0.441 \sparkspike 0.256 0.466 \sparkspike 0.282 0.146 \sparkspike 0.308 0.408 \sparkspike 0.333 0.261 \sparkspike 0.359 0.123 \sparkspike 0.385 0.312 \sparkspike 0.410 0.229 \sparkspike 0.436 0.592 \sparkspike 0.462 0.141 \sparkspike 0.487 0.224 \sparkspike 0.513 0.119 \sparkspike 0.538 0.376 \sparkspike 0.564 0.245 \sparkspike 0.590 0.231 \sparkspike 0.615 0.120 \sparkspike 0.641 0.156 \sparkspike 0.667 0.211 \sparkspike 0.692 0.177 \sparkspike 0.718 0.174 \sparkspike 0.744 0.148 \sparkspike 0.769 0.146 \sparkspike 0.795 0.164 \sparkspike 0.821 0.381 \sparkspike 0.846 0.158 \sparkspike 0.872 0.259 \sparkspike 0.897 0.305 \sparkspike 0.923 0.177 \sparkspike 0.949 0.085 \sparkspike 0.974 0.176 \sparkspike 1.000 0.657
\end{sparkline} & compensation wages pension wage salary welfare compensate salaries retirement bonus \\
\hline
\begin{sparkline}{10}
\sparkspike 0.000 0.475 \sparkspike 0.026 0.446 \sparkspike 0.051 0.936 \sparkspike 0.077 0.603 \sparkspike 0.103 0.715 \sparkspike 0.128 0.334 \sparkspike 0.154 1.000 \sparkspike 0.179 0.527 \sparkspike 0.205 0.322 \sparkspike 0.231 0.776 \sparkspike 0.256 0.350 \sparkspike 0.282 0.426 \sparkspike 0.308 0.415 \sparkspike 0.333 0.396 \sparkspike 0.359 0.344 \sparkspike 0.385 0.405 \sparkspike 0.410 0.241 \sparkspike 0.436 0.144 \sparkspike 0.462 0.362 \sparkspike 0.487 0.461 \sparkspike 0.513 0.240 \sparkspike 0.538 0.260 \sparkspike 0.564 0.188 \sparkspike 0.590 0.423 \sparkspike 0.615 0.250 \sparkspike 0.641 0.397 \sparkspike 0.667 0.203 \sparkspike 0.692 0.328 \sparkspike 0.718 0.380 \sparkspike 0.744 0.314 \sparkspike 0.769 0.344 \sparkspike 0.795 0.282 \sparkspike 0.821 0.174 \sparkspike 0.846 0.387 \sparkspike 0.872 0.238 \sparkspike 0.897 0.229 \sparkspike 0.923 0.181 \sparkspike 0.949 0.332 \sparkspike 0.974 0.265 \sparkspike 1.000 0.203
\end{sparkline} & discrimination prejudice unfair bias harassment segregation retaliation boycott persecution\\
\hline
\begin{sparkline}{10}
\sparkspike 0.000 0.508 \sparkspike 0.026 0.668 \sparkspike 0.051 0.832 \sparkspike 0.077 0.857 \sparkspike 0.103 0.401 \sparkspike 0.128 0.511 \sparkspike 0.154 0.723 \sparkspike 0.179 0.268 \sparkspike 0.205 0.567 \sparkspike 0.231 0.377 \sparkspike 0.256 0.903 \sparkspike 0.282 0.479 \sparkspike 0.308 0.498 \sparkspike 0.333 0.260 \sparkspike 0.359 0.382 \sparkspike 0.385 0.173 \sparkspike 0.410 0.176 \sparkspike 0.436 0.614 \sparkspike 0.462 0.306 \sparkspike 0.487 0.347 \sparkspike 0.513 0.753 \sparkspike 0.538 0.357 \sparkspike 0.564 0.326 \sparkspike 0.590 0.664 \sparkspike 0.615 0.114 \sparkspike 0.641 0.183 \sparkspike 0.667 0.433 \sparkspike 0.692 0.282 \sparkspike 0.718 0.214 \sparkspike 0.744 0.216 \sparkspike 0.769 0.154 \sparkspike 0.795 0.939 \sparkspike 0.821 1.000 \sparkspike 0.846 0.472 \sparkspike 0.872 0.375 \sparkspike 0.897 0.317 \sparkspike 0.923 0.926 \sparkspike 0.949 0.198 \sparkspike 0.974 0.255 \sparkspike 1.000 0.252
\end{sparkline} & medical health care hospital physician patient Medical physicians clinic hospitals \\
\hline
\begin{sparkline}{10}
\sparkspike 0.000 0.437 \sparkspike 0.026 0.360 \sparkspike 0.051 0.532 \sparkspike 0.077 0.606 \sparkspike 0.103 1.000 \sparkspike 0.128 0.434 \sparkspike 0.154 0.625 \sparkspike 0.179 0.317 \sparkspike 0.205 0.508 \sparkspike 0.231 0.260 \sparkspike 0.256 0.598 \sparkspike 0.282 0.244 \sparkspike 0.308 0.522 \sparkspike 0.333 0.538 \sparkspike 0.359 0.334 \sparkspike 0.385 0.153 \sparkspike 0.410 0.186 \sparkspike 0.436 0.544 \sparkspike 0.462 0.158 \sparkspike 0.487 0.219 \sparkspike 0.513 0.063 \sparkspike 0.538 0.114 \sparkspike 0.564 0.356 \sparkspike 0.590 0.122 \sparkspike 0.615 0.218 \sparkspike 0.641 0.223 \sparkspike 0.667 0.255 \sparkspike 0.692 0.494 \sparkspike 0.718 0.371 \sparkspike 0.744 0.094 \sparkspike 0.769 0.345 \sparkspike 0.795 0.318 \sparkspike 0.821 0.307 \sparkspike 0.846 0.587 \sparkspike 0.872 0.416 \sparkspike 0.897 0.304 \sparkspike 0.923 0.718 \sparkspike 0.949 0.079 \sparkspike 0.974 0.877 \sparkspike 1.000 0.168
\end{sparkline} & market competition competitive markets compete demand marketplace trading competitor trade \\
\hline
\begin{sparkline}{10}
\sparkspike 0.000 0.635 \sparkspike 0.026 0.367 \sparkspike 0.051 0.493 \sparkspike 0.077 0.471 \sparkspike 0.103 0.267 \sparkspike 0.128 0.156 \sparkspike 0.154 1.000 \sparkspike 0.179 0.141 \sparkspike 0.205 0.090 \sparkspike 0.231 0.182 \sparkspike 0.256 0.121 \sparkspike 0.282 0.386 \sparkspike 0.308 0.544 \sparkspike 0.333 0.192 \sparkspike 0.359 0.555 \sparkspike 0.385 0.346 \sparkspike 0.410 0.641 \sparkspike 0.436 0.475 \sparkspike 0.462 0.138 \sparkspike 0.487 0.266 \sparkspike 0.513 0.743 \sparkspike 0.538 0.152 \sparkspike 0.564 0.106 \sparkspike 0.590 0.594 \sparkspike 0.615 0.184 \sparkspike 0.641 0.154 \sparkspike 0.667 0.361 \sparkspike 0.692 0.187 \sparkspike 0.718 0.643 \sparkspike 0.744 0.296 \sparkspike 0.769 0.430 \sparkspike 0.795 0.135 \sparkspike 0.821 0.071 \sparkspike 0.846 0.337 \sparkspike 0.872 0.364 \sparkspike 0.897 0.411 \sparkspike 0.923 0.216 \sparkspike 0.949 0.130 \sparkspike 0.974 0.531 \sparkspike 1.000 0.215
\end{sparkline} & election vote voting electoral ballot voter elected votes elect Election \\

\hline
\begin{sparkline}{10}
\sparkspike 0.000 0.312 \sparkspike 0.026 0.138 \sparkspike 0.051 0.231 \sparkspike 0.077 0.403 \sparkspike 0.103 0.234 \sparkspike 0.128 0.345 \sparkspike 0.154 0.310 \sparkspike 0.179 0.206 \sparkspike 0.205 0.138 \sparkspike 0.231 0.130 \sparkspike 0.256 0.251 \sparkspike 0.282 0.190 \sparkspike 0.308 0.316 \sparkspike 0.333 0.400 \sparkspike 0.359 0.301 \sparkspike 0.385 0.288 \sparkspike 0.410 0.074 \sparkspike 0.436 0.197 \sparkspike 0.462 0.385 \sparkspike 0.487 0.176 \sparkspike 0.513 0.345 \sparkspike 0.538 0.138 \sparkspike 0.564 0.171 \sparkspike 0.590 0.203 \sparkspike 0.615 0.258 \sparkspike 0.641 0.291 \sparkspike 0.667 0.352 \sparkspike 0.692 0.275 \sparkspike 0.718 0.923 \sparkspike 0.744 0.415 \sparkspike 0.769 1.000 \sparkspike 0.795 0.542 \sparkspike 0.821 0.325 \sparkspike 0.846 0.224 \sparkspike 0.872 0.941 \sparkspike 0.897 0.408 \sparkspike 0.923 0.346 \sparkspike 0.949 0.229 \sparkspike 0.974 0.364 \sparkspike 1.000 0.654
\end{sparkline} & violence firearm gun violent weapon firearms armed weapons arms lethal \\
\hline
\begin{sparkline}{10}
\sparkspike 0.000 0.442 \sparkspike 0.026 0.168 \sparkspike 0.051 0.029 \sparkspike 0.077 0.050 \sparkspike 0.103 0.280 \sparkspike 0.128 0.323 \sparkspike 0.154 0.020 \sparkspike 0.179 0.046 \sparkspike 0.205 0.193 \sparkspike 0.231 0.156 \sparkspike 0.256 0.288 \sparkspike 0.282 0.085 \sparkspike 0.308 0.030 \sparkspike 0.333 0.082 \sparkspike 0.359 0.079 \sparkspike 0.385 0.041 \sparkspike 0.410 0.085 \sparkspike 0.436 0.079 \sparkspike 0.462 0.159 \sparkspike 0.487 0.184 \sparkspike 0.513 0.019 \sparkspike 0.538 0.210 \sparkspike 0.564 0.107 \sparkspike 0.590 0.417 \sparkspike 0.615 0.022 \sparkspike 0.641 0.101 \sparkspike 0.667 0.153 \sparkspike 0.692 0.197 \sparkspike 0.718 0.131 \sparkspike 0.744 0.012 \sparkspike 0.769 0.568 \sparkspike 0.795 0.239 \sparkspike 0.821 1.000 \sparkspike 0.846 0.905 \sparkspike 0.872 0.445 \sparkspike 0.897 0.318 \sparkspike 0.923 0.277 \sparkspike 0.949 0.473 \sparkspike 0.974 0.278 \sparkspike 1.000 0.381
\end{sparkline} & patent copyright Copyright Patent patents trademark invention patented copyrighted patentee \\
\hline
    \end{tabular}
    \caption{Grounded topics allow us to analyze trends in the organization of a corpus using a BERT topic model with $K=500$. Here we measure the prevalence of topics from 1980 to 2019 in US Supreme Court opinions by counting token assignments. Topics related to \textit{unions} and \textit{natural resources} are more prevalent earlier in the collection, while topics related to \textit{firearms} and \textit{intellectual property} have become more prominent.}
    \label{tab:scotustopics}
\end{table*}

\paragraph{Contextualized embedding clusters support collection analysis.}
Token-level clusterings of contextualized word representations support sophisticated corpus analysis with little additional complexity.
In practice, topic models are often used as a way to ``map'' a specific collection, for example by identifying key documents or measuring the prevalence of topics over time \cite{boyd2017applications}. 
A key disadvantage therefore of vocabulary-level semantic clustering is that it is not \textit{grounded} in specific documents in a collection.
Token-level clustering, in contrast, supports a wide range of analysis techniques for researchers interested in using a topic model as a means for studying a specific collection.
We provide two case studies that both use additional metadata to organize token-level clusterings \textit{post hoc}.

Given a partition of the working collection, we can count tokens assigned to each topic within a given partition to estimate locally prominent topics.
Table \ref{tab:reviewtopics} shows the three most prominent topics for the four product categories in \reviews{} from a \bert{} cluster model with $K=500$.
Many of these topics are clearly interpretable as aspects of a particular product (e.g. \textit{full albums}, \textit{individual songs}, \textit{descriptions of songs}).
Two topics contain mostly prepositions.
While we could have added these words to a stoplist before clustering, these less obviously interpretable clusters can nevertheless represent distinct discourses, such as descriptions of action in Movies (\textit{into, up, over, through, ...}) or descriptions of physical objects in Electronics (\textit{use, up, than, off, ...}).
We can also find the topics that are \textit{least} associated with any one product category by calculating the entropy of their distribution of tokens across categories.
These are shown in Table \ref{tab:eventopics}, and appear to represent subtle variations of subjective experiences: \textit{overkill}, \textit{possibilities}, \textit{reasons}, and \textit{time periods}.
We emphasize that this analysis requires no additional modeling, simply counting.

For partitions that have a natural order, such as years, we can create time series in the same \textit{post hoc} manner.
Thus, we can use a \bert{} clustering of \scotus{} to examine the changes in subject of the cases brought before the US Supreme Court. 
Table \ref{tab:scotustopics} shows time series for nine manually selected topics from a \bert{} clustering of \scotus{} with $K=500$, ordered by the means of their distributions over years. 
We find that topics related to \textit{labor and collective bargaining}, \textit{oil and gas exploration}, and \textit{compensation} have decreased in intensity since the 1980s, while those related to \textit{medical care} and \textit{elections} have remained relatively stable.
It appears that \textit{competitive markets} was a less common subject in the middle years, but has returned to prominence.
Meanwhile, \textit{discrimination} has remained a prominent topic throughout the period, but with higher intensities in the 1980s.
Additionally, topics related to \textit{gun violence} and \textit{patents and copyright} appear to be increasing in intensity.

\section{Conclusion}
We have presented a simple, reliable method for extracting mixed-membership models from pretrained contextualized word representations.
This result is of interest in several ways.
First, it provides insight into the affordances of contextualized representations.
For example, our result suggests a way to rationalize seemingly \textit{ad hoc} methods such as averaging token vectors to build a representation of a sentence.
Second, it suggests directions for further analysis and development of contextualized representation models and algorithms.
The significant differences we observe in superficially similar systems such as \bert{} and \roberta{} require explanations that could expand our theoretical understanding of what these models are doing.
Why, for example, is \roberta{} more prone to very small, specific clusters, while \bert{} is not?
Furthermore, if models like \bert{} are producing output similar to topic model algorithms, this connection may suggest new directions for simpler and more efficient language model algorithms, as well as more representationally powerful topic model algorithms.
Third, there may be substantial practical benefits for researchers analyzing collections.
Although running \bert{} on a large-scale corpus may for now be substantially more computationally inefficient than running highly-tuned LDA algorithms, passing a collection through such a system is likely to become an increasingly common analysis step.
If such practices could be combined with online clustering algorithms that would not require storing large numbers of dense token-level vectors, data analysts who are already using \bert{}-based workflows could easily extract high-quality topic model output with essentially no additional work.

\section*{Acknowledgments}
This work was supported by NSF \#1652536 and the Alfred P. Sloan foundation.

\bibliography{anthology,acl2020}
\bibliographystyle{acl_natbib}

\end{document}